\PassOptionsToPackage{table}{xcolor}
\RequirePackage{amsmath}
\PassOptionsToPackage{section}{placeins}
\documentclass{bytedance_seed}

\renewcommand{\authorformat}[2][]{\mbox{\authorfont\sffamily #2$^{#1}$}}
\usepackage{fvextra}
\usepackage{enumitem}
\usepackage{longtable}
\newcommand{\revision}[1]{{#1}}
\usepackage{amsmath,amsfonts,bm}

\def\eqref#1{equation~\ref{#1}}

\def\1{\bm{1}}

\DeclareMathAlphabet{\mathsfit}{\encodingdefault}{\sfdefault}{m}{sl}
\SetMathAlphabet{\mathsfit}{bold}{\encodingdefault}{\sfdefault}{bx}{n}

\usepackage{dblfloatfix}
\usepackage{minted}
\usepackage{xspace}
\usepackage{amsmath,amssymb}
\usepackage{makecell}

\newmintinline{python}{python3}
\newminted{python}{frame=single,framesep=1mm,fontsize=\scriptsize,python3,linenos}
\newmintinline{md}{markdown}

\newminted{mdcode}{
    style=colorful,
    frame=single,
    framesep=1mm,
    fontsize=\scriptsize,
    linenos,
    breaklines,
    tabsize=2
}

\newcommand{\ours}{\textsc{RSIBench-Infra}\xspace}

\usepackage{tikz}
\usetikzlibrary{calc}
\usepackage{xcolor}

\definecolor{gensicolor}{RGB}{220,245,84}  
\definecolor{seedcolor}{RGB}{120,230,220}  

\patchcmd{\mymaketitle}{\affiliationlist\par}{\authorlist\par\vskip 3mm\affiliationlist\par}{}{\ClassError{bytedance_seed}{Cannot insert authors}{Check the title implementation}}
\newcommand{\projectpage}{\url{https://rsibench-infra.github.io}}
\newcommand{\correspondingemail}{\href{mailto:zhouhao@air.tsinghua.edu.cn}{\texttt{zhouhao@air.tsinghua.edu.cn}}}
\title{RSIBench-Infra: Can Agents Optimize Training and Inference Infrastructure at Production Scale?}
\author[1,2,3,*]{Hanlin Wu}
\author[2,3,*]{Yifei Wang}
\author[1,2,3]{Zhicheng Cai}
\author[1]{Wenxuan Tan}
\author[1]{Jiahao Li}
\author[2,3]{Huan-ang Gao}
\author[2,3]{Wei-Ying Ma}
\author[2,3]{Ya-Qin Zhang}
\author[2,3]{Jingjing Liu}
\author[2,3,\dagger]{Hao Zhou}
\affiliation[1]{ByteDance Seed}
\affiliation[2]{Institute for AI Industry Research (AIR), Tsinghua University}
\affiliation[3]{SIA-Lab of Tsinghua AIR and ByteDance Seed}
\contribution[*]{Equal contributions}
\contribution[\dagger]{Corresponding author: \correspondingemail}
\checkdata[Project page]{\projectpage}
\abstract{
Improving the efficiency of training and inference infrastructure is a practical
component of recursive self-improvement. Existing benchmarks for agent-driven
optimization largely focus on individual software layers such as kernel, leaving
unclear whether agents can deliver end-to-end performance gains under production
workloads. We introduce \ours, a benchmark of 20 training and
inference tasks spanning diverse model families and modalities, with deployments
of up to 32 accelerators on NVIDIA GPUs and Ascend NPUs. Agents start from
runnable baselines derived from official recipes and can modify the software stack
across engine, kernel, and instruction levels. An OCI image-layer submission
protocol captures these changes for reproducible deployment, while independent
evaluation measures end-to-end performance under task-specific quality constraints.
\revision{Across all task results, 66.2\% exceed $1.10\times$ validated speedup. }
These results demonstrate that current agents can improve production-scale training 
and inference performance, but identifying effective optimization
strategies and consistently translating local improvements into end-to-end gains
remain challenging.
}

\begin{document}
\maketitle

\section{Introduction}
\label{sec:introduction}

\suppressfloats[t]
\begin{figure}[t]
    \centering
    \includegraphics[width=\textwidth]{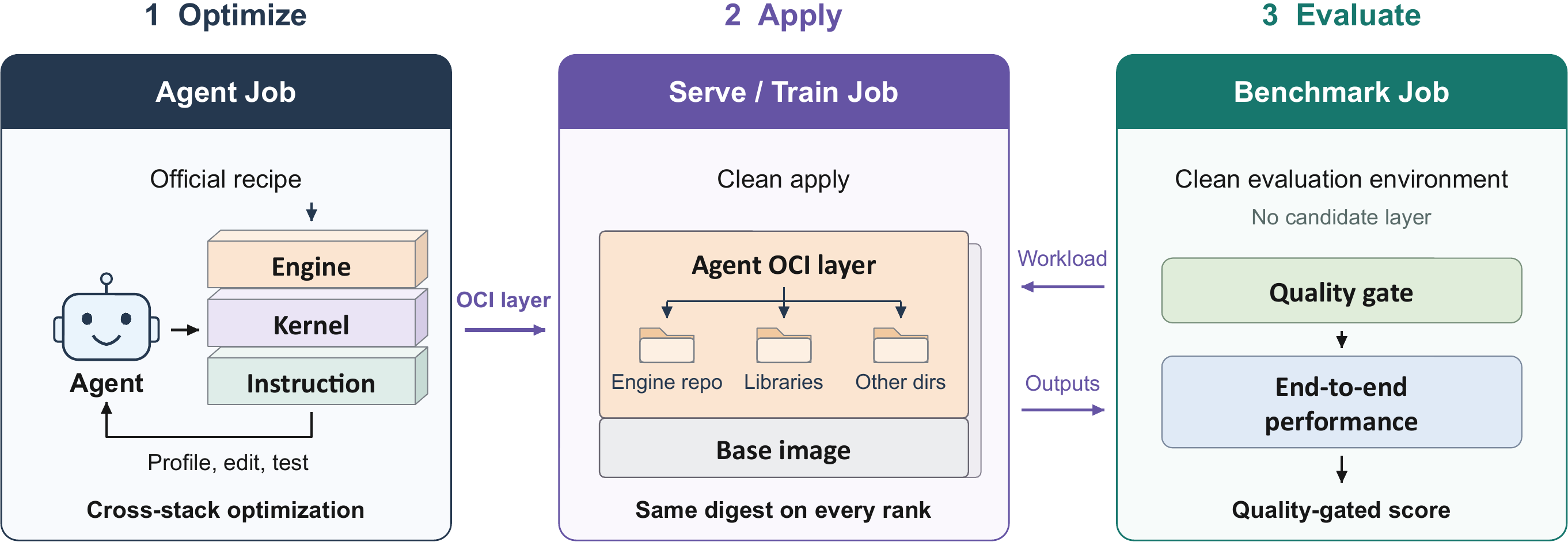}
    \caption{\textbf{Overview of \ours.} Starting from an official recipe, an agent optimizes the software stack at the engine, kernel, and instruction levels and submits its changes as a single OCI image layer. The agent OCI layer captures changes across the engine repository, libraries, and other directories and is applied to the base environment of an independent serving or training job, using the same layer digest on every rank. A clean benchmark job, which never loads the candidate layer, checks task-level quality and measures end-to-end performance to produce a quality-gated score.}
    \label{fig:rsibench-overview}
    \vskip -0.1in
\end{figure}

Worldwide AI infrastructure spending is projected to reach \$487 billion in 2026 as estimated \citep{idc2026aiinfrastructure}.
However, widely used LLM serving engines achieve about 22--38\% of their theoretically optimal
end-to-end throughput \citep{zhu2025nanoflow}. 
Furthermore, given the limited production capacity for GPU accelerators, 
this gap also constrains the development of recursive self-improvement (RSI), 
whose iterative loop requires generating rollouts as well as updating model parameters \citep{xia2025agent0}.
Therefore, the efficiency of training and inference software stack plays a crucial role in transforming the huge investment into real intelligence.
Due to the vast complexity of the modern machine learning infrastructure software, 
agents equipped with broad knowledge and the capacity to work continuously hold considerable promise for optimizing training and inference performance.

Recent works have begun to study this possibility at different abstraction levels of machine learning infra software: 
1) GPU Kernel Level: KernelBench~\citep{ouyang2025kernelbench}, FlashInfer-Bench~\citep{xing2026flashinferbench}, 
and SOL-ExecBench~\citep{lin2026solexecbench} evaluate whether language models can generate correct and
fast kernels for diverse machine-learning workloads. FlashInfer-Bench moves closer to deployment by
deriving tasks from real serving traces and integrating validated kernels into production LLM engines. 
2) Engine Level: VibeServe~\citep{kamahori2026vibeserve} and InferenceBench~\citep{yeon2026inferencebench}
explore the question whether agents can write LLM inference
servers under a fixed time budget.
3) Instruction and Close-to-Metal Level:
PTXBench~\citep{ptxbench2026} evaluates whether LLMs can generate efficient
CUDA kernels with architecture-specific inline PTX, measuring functional
correctness, runtime execution of required instructions, and performance
relative to optimized libraries.
KernelCraft~\citep{nie2026kernelcraft} evaluates agents on generating and
optimizing low-level kernels for emerging accelerators with specialized
instruction set architectures (ISAs), using execution feedback to assess
correctness and performance.

However, evaluations confined to individual software layers cannot fully characterize end-to-end optimization,
because system performance depends on interactions across multiple layers of the software stack, 
and optimizations at one layer do not necessarily yield an end-to-end global optimum. 
For example, CAKE~\citep{ye2026cake} enables agents to co-evolve GPU kernels
and compiler infrastructure, including the intermediate representation,
lowering rules, and correctness and performance analyses. Agents can thus
extend the compiler capabilities needed to express efficient hardware
schedules while optimizing the kernels that use them. Its generated Kimi
Delta Attention kernels achieve a $2.05\times$ geometric-mean speedup over
FlashKDA on six tested shapes.

To address these challenges, we introduce \ours, a benchmark for evaluating
whether agents can effectively optimize end-to-end training and inference
performance in production-oriented settings (Figure~\ref{fig:rsibench-overview}). A key distinction from prior benchmarks is that \ours allows agents to modify any part of the software stack in the sandbox while it also makes the traditional git patch based evaluation not enough. We therefore introduce an OCI image layer--based submission protocol\footnote{Open Container Initiative, Image Specification v1.1.1, Filesystem Layers: \url{https://github.com/opencontainers/image-spec/blob/v1.1.1/layer.md}.},
through which agents submit their modifications as a single image layer.

For inference tasks, we use the official
model-specific deployment cookbooks and recipes provided by widely used serving engines,
including vLLM, SGLang, and TensorRT-LLM, as baseline implementations. Agents are required to optimize these implementations, 
and provide optimized service launch scripts. Independent evaluation uses task-specific task-level quality gates to assess whether performance improvements preserve
task-level quality \emph{e.g.} LongBench v2 for long context text generation. Our workloads span a broad range of computational
workloads and application domains including: text-only models
(e.g., GLM-5.3~\citep{glm5team2026glm5}), multimodal understanding
(e.g., Qwen3-VL~\citep{bai2025qwen3vl}), video generation
(e.g., Wan 2.2~\citep{wan2025}), text diffusion
(e.g., LLaDA2.1-mini~\citep{bie2026llada21}), world modeling
(e.g., LingBot World 2.0~\citep{gao2026lingbotworld2}), and
vision-language-action models
(e.g., OpenPI's $\pi_{0.5}$~\citep{physicalintelligence2025pi05}).

Our main contributions are:
\begin{itemize}
    \item We introduce \ours, the first-known benchmark for agent-driven optimization of
    end-to-end training and inference performance at production scale through
    cross-stack software modifications under task-specific constraints.
    These improvements can translate directly into higher serving capacity or
    lower compute cost per token in production deployments with comparable
    workloads and operating conditions.

    \item \ours defines and benchmarks a key component of RSI:
the ability of AI systems to accelerate their own inference and training
by optimizing the underlying software stack. Faster rollout generation
and model updates can shorten self-improvement cycles and enable more
iterations within the same compute budget. 

    \item We evaluate seven agent--model configurations across 20 production-oriented inference and
training tasks. By analyzing
submitted artifacts, we characterize successful optimization strategies across the software
stack and identify recurring failures in completing optimization runs, packaging submissions,
and preserving verifier interfaces. These findings help the community understand the
capabilities and current limits of existing models and inform the development of next-generation models.

    \item \revision{We provide analytical SOL references and bounded reference-normalized scores for all 20 tasks, together with a five-task single-H20 Lite Set for accessible evaluation.}

\end{itemize}

\section{Related Works}
\label{sec:related-work}

\subsection{LLMs for Infrastructure Optimization}
Existing benchmarks evaluate LLM-driven performance optimization at different levels of the software stack, ranging from low-level GPU kernels to serving optimization: KernelBench and TritonBench evaluate kernel correctness and
performance~\citep{ouyang2025kernelbench,li2025tritonbench}, while PTXBench and
KernelCraft target architecture-specific instructions and emerging
accelerators~\citep{ptxbench2026,nie2026kernelcraft}.
FlashInfer-Bench connects serving-trace workloads to kernel
integration~\citep{xing2026flashinferbench}, and SOL-ExecBench evaluates kernels
against hardware performance bounds~\citep{lin2026solexecbench}.
At the engine level, InferenceBench evaluates open-ended serving
optimization~\citep{yeon2026inferencebench}, while VibeServe synthesizes
runtimes specialized to models, hardware, and workloads~\citep{kamahori2026vibeserve}.
\ours evaluates both training and inference optimization in production
settings, including distributed deployments, using official recipe baselines,
quality-gated performance measurements, and OCI layers that capture changes
across repositories and installation directories.

\subsection{Recursive Self-Improvement}
Recursive self-improvement (RSI) offers a promising path toward increasingly capable AI systems and has attracted growing research interest recently. STOP and the Darwin G\"odel Machine improve agent
code~\citep{zelikman2023stop,zhang2025dgm}, whereas Agent0 co-evolves task
curricula and executor capabilities~\citep{xia2025agent0}.
RSIBench-Data evaluates training-data research through checkpoint feedback
under fixed post-training infrastructure~\citep{meng2026rsibenchdata}.
RSI-Exam evaluates improvements to executable methods and frozen-model
harnesses through isolated re-execution on hidden
instances~\citep{rsiexam2026}.
\ours complements these efforts by making training and inference
infrastructure an explicit target of agent-driven improvement within RSI.
Using fixed workloads and task-level quality gates, it isolates agents'
ability to improve end-to-end systems efficiency, providing an infrastructure
counterpart to benchmarks of data-centric research and agent adaptation.

\section{RSIBench-Infra}
\label{sec:benchmark-design}

\ours casts infrastructure optimization as a task of improving end-to-end
performance while preserving a prescribed workload and its quality requirements.
Each task pairs a reproducible reference implementation with a hardware
configuration, an execution contract, and an independent verifier.

\subsection{Benchmark Construction}
\label{sec:baseline-preparation}
\label{sec:task-sourcing-construction}
\begin{figure}[!htbp]
    \centering
    \includegraphics[width=\textwidth]{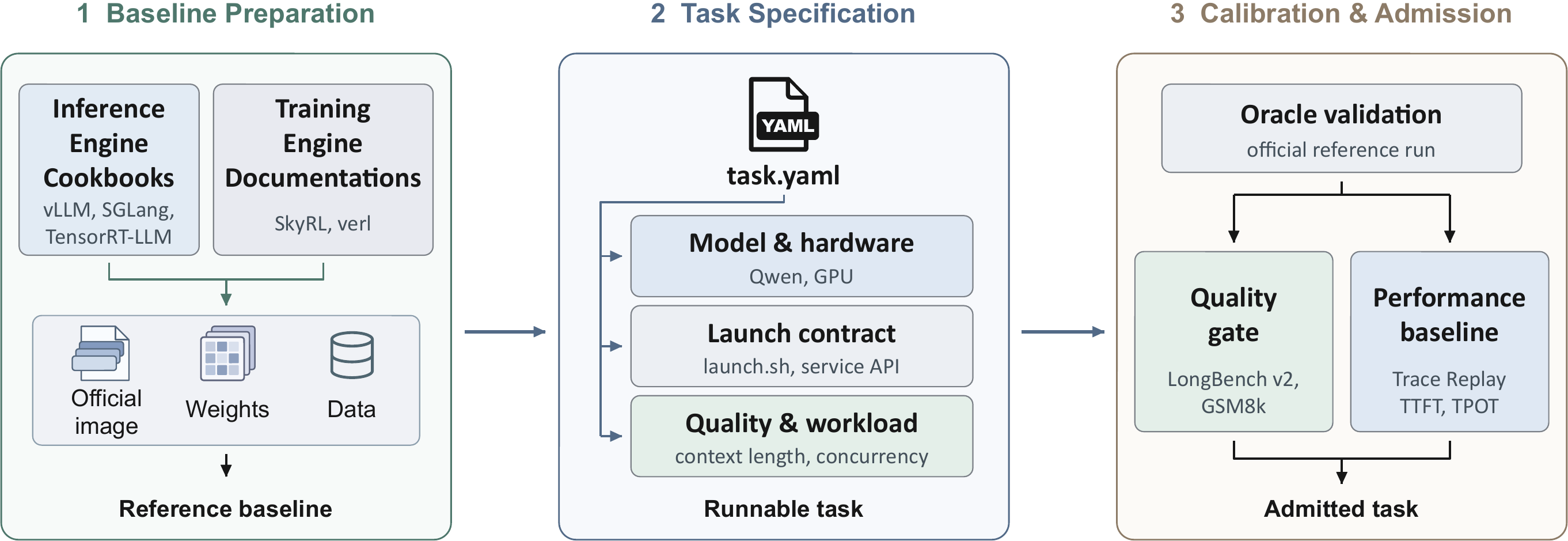}
    \caption{\textbf{Benchmark construction workflow.} Official inference engine
    cookbooks or training engine documentation provide the starting point for
    preparing the reference environment, model weights, and data. A task
    specification binds these assets to the hardware, launch contract, quality
    requirements, and performance workload. The quality benchmarks and latency
    metrics shown are inference examples; training tasks use numerical training
    signals and end-to-end execution time.}
    \label{fig:benchmark-construction}
    \vskip -0.1in
\end{figure}

Figure~\ref{fig:benchmark-construction} summarizes how we turn official
recipes into executable tasks with fixed quality requirements and measured
performance baselines.

\paragraph{Task Sourcing.}
Inference tasks follow model-specific cookbooks and deployment recipes from
widely used engines such as vLLM, SGLang, and TensorRT-LLM. Selected open-source models include
Inkling~\citep{thinkingmachines2026inkling} and Kimi K3~\citep{kimiteam2026kimik3},
spanning 1.7B (Qwen3-TTS) to 2.8T parameters (Kimi K3) and dense and
mixture-of-experts architectures. Workloads cover autoregressive generation, text diffusion,
vision--language understanding, video and speech generation, and
vision--language--action inference to exercise different serving execution paths.
Training tasks follow pinned verl recipes and Tinker-compatible reference
implementations, executed through SkyRL-Tinker or verl-tinker. They cover
DAPO-style text GRPO, supervised fine-tuning, vision--language GRPO, and on-policy distillation.

\paragraph{Task Admission.}
\label{sec:task-construction}
An executable reference \texttt{launch.sh} binds each recipe's container
image, checkpoint, precision, and parallelism settings to the task hardware.
Admission verifies the required interfaces and task-specific behavior against
fixed evaluation assets. Language and vision--language services use GSM8K,
LongBench v2, or MMMU, while generation tasks use reference-output or speech
quality checks. Training admission uses frozen-input replay, reference
signals, and positive and negative controls. These checks establish a
reproducible starting point; the admitted baseline is not assumed to be an
optimal implementation. Exact quality criteria are given in
Appendix~\ref{app:quality-gates}.

\paragraph{Performance Baseline.}
\label{sec:task-specification}
For each admitted implementation, we establish a performance baseline using
end-to-end workloads representative of production deployments. For example,
the Kimi K3 task serves a 256k-context workload on four nodes with eight
GPUs each. We construct its agentic request trace from Terminal-Bench 2.1
trajectories~\citep{merrill2026terminalbench}, comprising 894 requests across
20 sessions, and replay it in the Mooncake trace format~\citep{qin2025mooncake},
reconstructing the input and output lengths and session-prefix reuse patterns.

This production-oriented workload exercises long-context prefill, decoding, KV-cache management,
and distributed communication within the complete serving stack. We measure
output throughput over a fixed one-hour window at concurrency one and record
latency diagnostics including time to first token (TTFT) and time per output
token (TPOT). At fixed hardware and validated model quality, throughput
quantifies generated tokens per GPU-hour. We apply the same principle to other tasks,
measuring complete ML workloads of each training task. These
measurements provide the reference values for scoring agent optimizations
in the later Section~\ref{sec:scoring}.


\subsection{Task Definition}
\label{sec:task-definition}
Figure~\ref{fig:rsibench-overview} illustrates the overall optimization and
evaluation workflow. Our abstraction of a task follows
Harbor~\citep{Harbor_Framework}.
Each task produced by the construction process is defined as a tuple
$\mathcal{T} = (\mathcal{I}, \mathcal{E}, \mathcal{V}) $
with three components:
\begin{itemize}
    \item \textbf{Instruction $\mathcal{I}$} contains the task description and
    supporting information available to the agent, including the optimization
    objective, baseline setup, operational constraints, and submission requirements.
    The instruction includes a throughput optimization objective and
    requirements to preserve model weights, maintain the specified
    interfaces, and support the task's context length.
    \item \textbf{Environment $\mathcal{E}$} contains a runnable baseline derived
    from an official recipe, the required assets, and the execution resource
    configuration. The environment includes the serving engine, checkpoint, and launch
    configuration for the assigned hardware.
    \item \textbf{Verifier $\mathcal{V}$} contains the evaluation code and assets
    for submission validation, quality assessment, and performance measurement.
    The verifier validates and applies the submitted OCI layer to fresh
    baseline environments, runs the task's quality gate, and measures the
    prescribed workload under a fixed protocol.
\end{itemize}

\subsection{Evaluation Procedure}
\label{sec:evaluation-protocol}
Figure~\ref{fig:rsibench-overview} summarizes three stages: optimization,
deployment, and evaluation. The agent optimizes an official-recipe baseline
and submits an OCI layer. An independent Serve/Train Job applies the same
layer to clean environments on all ranks; a separate Benchmark Job checks
quality and measures end-to-end performance.

\paragraph{Agent Optimization.}
\label{sec:oci-submission}
Starting from the base environment, the agent profiles, edits, and tests the
software. Inference tasks include the \texttt{e2e-infer-opt} skill: establish
a reproducible baseline and quality checks, classify bottlenecks, trace hotspots
to implementations, and build isolated microbenchmarks. Targeted optimizations
are validated with microbenchmarks and end-to-end quality and performance tests,
repeating as bottlenecks shift. Task-specific engine and hardware guidance
supports this workflow.

The agent is required to explicitly package its changes
as an OCI filesystem layer, \texttt{submit\_oci.tar.gz}, including the
required \texttt{launch.sh} to launch an optimized service. This submission may contain source code,
dependencies, and compiled artifacts within the permitted modification scope.
An orchestration script launches two independent jobs: a \verb|Serve Job| for
optimized inference or training and a \verb|Benchmark Job| for evaluation.
Both jobs run in environments isolated from the Agent Job, preventing
agent-side filesystem changes from directly modifying the evaluator's scripts
or scoring logic.

\paragraph{Serve/Train Job.}
\label{sec:job-isolation}
The Serve/Train Job starts from the same baseline
image as the Agent Job and receives the submitted OCI layer as the sole
artifact carrying agent modifications. The script verifies the engine baseline
and layer digests before applying the layer, using the same layer on every
rank. This reconstructs the submitted changes on a known baseline without
inheriting the Agent Job's runtime state. 

\paragraph{Benchmark Job.} The Benchmark Job never loads the submitted layer. It runs predefined quality
and performance scripts against the service through its exposed interfaces,
keeping workload generation, measurement, and scoring isolated from the
agent-modified software. This separation supports auditing for benchmark hacking.

\begin{table}[!htbp]
\vskip -0.2in
\centering
\caption{\textbf{Task overview.} Requests denotes the source trace size for
text replay, serving requests or chunks for other inference tasks, and
samples per workload step for training tasks (prompt
batch $\times$ rollout group in RL; batch size in SFT). A workload step
may contain multiple optimizer updates (Appendix~\ref{app:training-tasks}).}
\label{tab:task-overview}
\begingroup
\newcommand{\tasklogo}[1]{\raisebox{-1.2pt}{\includegraphics[width=9pt,height=8pt,keepaspectratio]{figures/logos/#1}}\,}
\newcommand{\taskengine}[1]{\addlinespace[3pt]\rowcolor{gray!18}\multicolumn{5}{l}{\textbf{#1}}\\[1pt]}
\setlength{\tabcolsep}{3pt}
\renewcommand{\arraystretch}{1.10}
\fontsize{8}{9.6}\selectfont
\begin{tabular*}{\linewidth}{@{\extracolsep{\fill}}lllll@{}}
\toprule
\textbf{Model} & \textbf{Devices} & \textbf{Workload} & \textbf{Requests} & \textbf{Quality gate} \\
\midrule
\taskengine{SGLang (7)}
\tasklogo{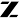}GLM-5.2 744B & $2\!\times\!16$ NPU & agentic-trace-256k & 894 & GSM8K \\
\tasklogo{zai.pdf}GLM-5.3 744B & $2\!\times\!8$ GPU & agentic-trace-256k & 894 & LongBench v2 \\
\tasklogo{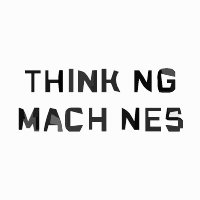}Inkling 975B & $1\!\times\!8$ GPU & agentic-trace-256k & 894 & LongBench v2 \\
\tasklogo{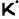}Kimi K3 2.8T & $4\!\times\!8$ GPU & agentic-trace-256k & 894 & LongBench v2 \\
\tasklogo{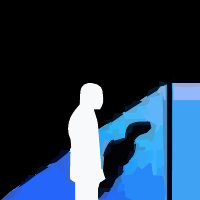}LLaDA2.1-mini 16B & $1\!\times\!1$ GPU & agentic-trace-32k & 140 & LongBench v2 \\
\tasklogo{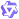}Qwen3.8 27B & $1\!\times\!1$ GPU & agentic-trace-256k & 894 & LongBench v2 \\
\tasklogo{qwen.pdf}Qwen3-VL 32B & $1\!\times\!1$ GPU & synthetic 1080p JPEGs & 32 & MMMU \\
\taskengine{SGLang Diffusion (2)}
\tasklogo{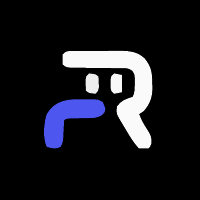}LingBot-World 2.0 14B & $1\!\times\!8$ GPU & fixed trace, 832$\times$480, 16 fps & 16 & Reference frames \\
\tasklogo{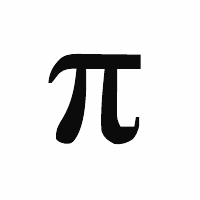}$\pi_{0.5}$-LIBERO 3.3B & $1\!\times\!1$ GPU & LIBERO observations & 16$\times$3 & Reference actions \\
\taskengine{vLLM (3)}
\tasklogo{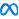}Muse-Glimmer 30B & $1\!\times\!1$ GPU & agentic-trace-128k & 936 & LongBench v2 \\
\tasklogo{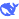}DeepSeek V4 Flash 284B & $1\!\times\!8$ NPU & agentic-trace-64k & 193 & LongBench v2 \\
\tasklogo{zai.pdf}GLM-5.3-Flash 320B & $2\!\times\!8$ NPU & agentic-trace-256k & 894 & GSM8K \\
\taskengine{vLLM-Omni (3)}
\tasklogo{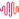}MiniMax-H3 64B & $1\!\times\!2$ GPU & 672$\times$384, 107 frames & 1 & Video + audio ref. \\
\tasklogo{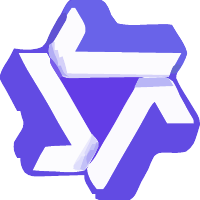}Wan2.2 TI2V 5B & $1\!\times\!1$ GPU & 1280$\times$704, 121 frames & 4 & Reference videos \\
\tasklogo{qwen.pdf}Qwen3-TTS Base 1.7B & $1\!\times\!1$ GPU & Seed-TTS eval (Chinese) & 202 & Chinese ASR WER \\
\taskengine{TensorRT-LLM (1)}
\tasklogo{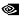}Nemotron 3 Ultra 550B & $1\!\times\!8$ GPU & agentic-trace-256k & 894 & LongBench v2 \\
\midrule
\taskengine{SkyRL (2)}
\tasklogo{qwen.pdf}Qwen3-30B-A3B-Base (GRPO) & $1\!\times\!8$ GPU & DAPO-Math-17K & 512$\times$8 & Loss + grad. norm \\
MiMo-7B-RL (MTP SFT) & $1\!\times\!8$ GPU & GSM8K & 64 & Loss + grad. norm \\
\taskengine{VeRL (2)}
\tasklogo{qwen.pdf}Qwen3-8B \& 32B (OPD) & $2\!\times\!8$ GPU & GSM8K+MATH & 128 & Loss + grad. norm \\
\tasklogo{qwen.pdf}Qwen3-VL-8B (GRPO) & $1\!\times\!8$ GPU & Geo3K train & 512$\times$5 & PPO/KL + forward \\
\bottomrule
\end{tabular*}
\par\vspace{-10pt}
\endgroup
\end{table}

\paragraph{Quality gates.}
\label{sec:quality-gates}
We accept performance improvements only when the submission passes the
quality gate for its task. Predefined scripts in the Benchmark Job use fixed inputs and
reference assets to assess language understanding with LongBench v2 or GSM8K,
vision--language understanding with MMMU, reference-output consistency for
video, world-model, and action generation, and speech intelligibility through
word error rate. For training, we replay frozen inputs and compare stepwise
loss and global gradient-norm curves with the reference execution to check
preservation of training behavior. The vision--language GRPO task additionally
checks policy and KL losses, actor and reference forward outputs, and an
isolated KL gradient probe. An \textbf{LLM judge} additionally
reviews agent trajectories and submitted changes to identify potential benchmark
hacking, including workload-specific shortcuts and violations of task constraints.
Task-specific inputs, acceptance criteria,
numerical tolerances, and the scope of these checks are detailed in
Appendix~\ref{app:quality-gates}.

\paragraph{End-to-end metrics.}
\label{sec:scoring}
We measure end-to-end throughput in the unit appropriate to each workload,
such as tokens/s for language serving, videos/hour for video generation,
frames/s for world modeling, and steps/hour for training. For task $i$, let
$P_i^{\mathrm{agent}}$ and $P_i^{\mathrm{base}}$ denote the submission's
throughput and the fixed official-recipe baseline throughput under the same
workload and measurement protocol. If the submission passes validation and
the task's quality gate and completes the prescribed performance measurement,
its validated speedup is $s_i=P_i^{\mathrm{agent}}/P_i^{\mathrm{base}}$.
Evaluated failures receive $s_i=0$; valid slowdowns retain their ratios below
one. For configuration $m$, let $\mathcal{E}_m$ be its eligible task set and
$N_m=|\mathcal{E}_m|$. We report \textbf{Faster Rate} and
\textbf{Median Speedup} on the same set:
\begin{equation}
 \begin{aligned}
 \text{Faster Rate}_m &= \frac{100}{N_m}\sum_{i\in\mathcal{E}_m}\mathbf{1}\{s_i>1.10\}\quad(\%),\\
 \text{Median Speedup}_m &= \operatorname{median}\{s_i:i\in\mathcal{E}_m\}.
 \end{aligned}
 \label{eq:aggregate-score}
\end{equation}
\revision{The default set contains all 20 tasks. Claude Opus~5.5 has formal
scores on 10 tasks ($N_m=10$); its remaining tasks encountered API safety-policy
errors and are excluded from all aggregates. The other six configurations use $N_m=20$. Every aggregate
uses the stated eligible-task denominator.}
Faster Rate counts improvements strictly above 10\% to reduce sensitivity to
small performance fluctuations. Median Speedup limits the influence of
exceptionally large gains. 

Alongside the benchmark scores, we retain other task-specific quality measurements and gate outcomes, request success rates, and raw performance ratios such as the median and 99th percentile of time to first token (TTFT).

\subsection{Dataset Overview}
\label{sec:coverage}
The suite comprises 16 inference tasks and 4 training tasks
(Table~\ref{tab:task-overview}), spanning seven engine groups and deployments
of one to 32 accelerators across up to four nodes on GPUs and
NPUs. Inference models range from 1.7B to 2.8T parameters.

\paragraph{Inference.}
The suite covers language serving, including GLM-5.3, Inkling, and
Kimi K3~\citep{glm5team2026glm5,thinkingmachines2026inkling,kimiteam2026kimik3},
vision--language understanding~\citep{bai2025qwen3vl}, text
diffusion~\citep{bie2026llada21}, video generation~\citep{wan2025}, speech
synthesis, world modeling~\citep{gao2026lingbotworld2}, and action
generation~\citep{physicalintelligence2025pi05}.
Long-context serving stresses prefill, decoding, cache management, and
distributed communication; multimodal tasks additionally exercise image
processing, iterative denoising, and action generation.

\paragraph{Training.}
The four tasks use full-parameter BF16 training: Qwen3-30B-A3B-Base
DAPO-style GRPO, MiMo-7B-RL MTP SFT, Qwen3-8B distillation from a frozen
Qwen3-32B teacher, and Qwen3-VL-8B-Instruct GRPO with PPO clipping and a KL
penalty. A trusted CPU client drives the services through the Tinker API,
controlling data, seeds, reward and advantage construction, update order, and
timing. The client reconstructs gate losses from returned forward outputs;
model forward/backward computation and optimizer updates execute in the
Train Job. Performance counts complete workload steps, including rollout,
weight synchronization, old-policy or reference forward passes, and teacher
scoring where required. Appendix~\ref{app:training-tasks} specifies each
objective and distinguishes workload steps from optimizer updates.

\section{Experiment}
\label{sec:experimental-setup}
\subsection{Setup}
We evaluate the seven agent--model configurations in
Table~\ref{tab:main-results}: GPT-6 Astra, Claude Opus~5.5,
DeepSeek V4 Flash, GLM-5.3, Qwen3.8-Max, Seed Evolving, and Kimi K3.
\revision{All experiments use \texttt{reasoning=max} with a 1M-token agent
context limit. GPT-6 Astra uses Codex, Claude Opus~5.5 uses Claude Code,
and the other five models use OpenCode.}
Each run has an eight-hour agent budget. Agents receive task
instructions and the runnable task baseline. Hardware, workload, and quality
requirements are fixed per task.

\subsection{Results and Analysis}
\label{sec:results}

\begin{table}[!htbp]
\centering
\caption{Model--harness results with an eight-hour budget. Faster counts validated speedups strictly above $1.10\times$. Mean SOL uses all eligible tasks (20, or 10 for Claude); Lite Mean uses five tasks, or three for Claude. Formal zeros are included. Tokens include all recorded input (including cache hits) and output, with separately recorded reasoning counted once. Mean time measures agent execution only. Bold marks the best complete-suite result in each performance column.}
\label{tab:main-results}
\begingroup\fontsize{8}{9.6}\selectfont
\setlength{\tabcolsep}{3pt}
\begin{tabular*}{\linewidth}{@{\extracolsep{\fill}}llrrrrrrr@{}}
\toprule
Model & Harness & Faster & Rate & \makecell{Median\\speedup} & \makecell{Tokens\\(M)} & \makecell{Mean time\\(h:mm)} & \makecell{Mean\\SOL} & \makecell{Lite\\Mean} \\\midrule
GPT-6 Astra & Codex & 16/20 & \textbf{80\%} & $\mathbf{1.3335}\times$ & 493.12 & 3:36 & \textbf{0.667} & 0.641 \\
Claude Opus 5.5 & Claude Code & 8/10 & 80\% & $1.3431\times$ & 1,135.86 & 3:25 & 0.711$^{\dagger}$ & 0.798$^{\ddagger}$ \\
GLM-5.3 & OpenCode & 14/20 & 70\% & $1.2772\times$ & 1,231.40 & 5:26 & 0.626 & 0.643 \\
Qwen3.8-Max & OpenCode & 14/20 & 70\% & $1.2107\times$ & 835.37 & 6:53 & 0.577 & \textbf{0.670} \\
DeepSeek V4 Flash & OpenCode & 12/20 & 60\% & $1.2846\times$ & 947.11 & 6:29 & 0.614 & 0.508 \\
Kimi K3 & OpenCode & 12/20 & 60\% & $1.1277\times$ & 755.89 & 5:09 & 0.627 & 0.597 \\
Seed Evolving & OpenCode & 10/20 & 50\% & $1.0933\times$ & 719.61 & 5:42 & 0.570 & 0.602 \\
\bottomrule
\end{tabular*}
\par\smallskip\raggedright\scriptsize
$^{\dagger}$Claude: 10 formally scored tasks; remaining tasks encountered API safety-policy errors.
$^{\ddagger}$Claude Lite: LLaDA2.1-mini, Qwen3-VL, and Wan (3/5 tasks).
\endgroup
\end{table}
\FloatBarrier

\paragraph{Main results.}
\revision{Table~\ref{tab:main-results} reports 130 formally scored results,
of which 86 (66.2\%) exceed $1.10\times$ validated speedup. Among the six
complete 20-task configurations, GPT-6 Astra leads Faster Rate (16/20, 80\%),
Median Speedup ($1.3335\times$), and mean SOL score (0.667).
GLM-5.3 and Qwen3.8-Max each improve 14/20 tasks by more than 10\%,
followed by DeepSeek V4 Flash and Kimi K3 at 12/20, and Seed Evolving at
10/20. }

\revision{Claude Opus~5.5 achieves 8/10 Faster results, a Median Speedup of
$1.3431\times$, and a mean SOL score of 0.711 on its formally scored subset.
Its coverage is shown explicitly throughout the tables and figures.
The scores indicate strong performance on the evaluated tasks.}

\paragraph{Execution time.}
\revision{Claude and Astra have the shortest mean agent execution times,
3\,h 25\,min and 3\,h 36\,min, respectively. Both attain an 80\% Faster
Rate on their respective evaluated sets while using approximately 43\%
and 45\% of the eight-hour budget. The remaining configurations average
5\,h 09\,min to 6\,h 53\,min. These measurements show that strong
optimization outcomes can be obtained well within the allocated budget.}

\FloatBarrier
\begin{figure}[!htbp]
    \centering
    \includegraphics[width=0.94\linewidth]{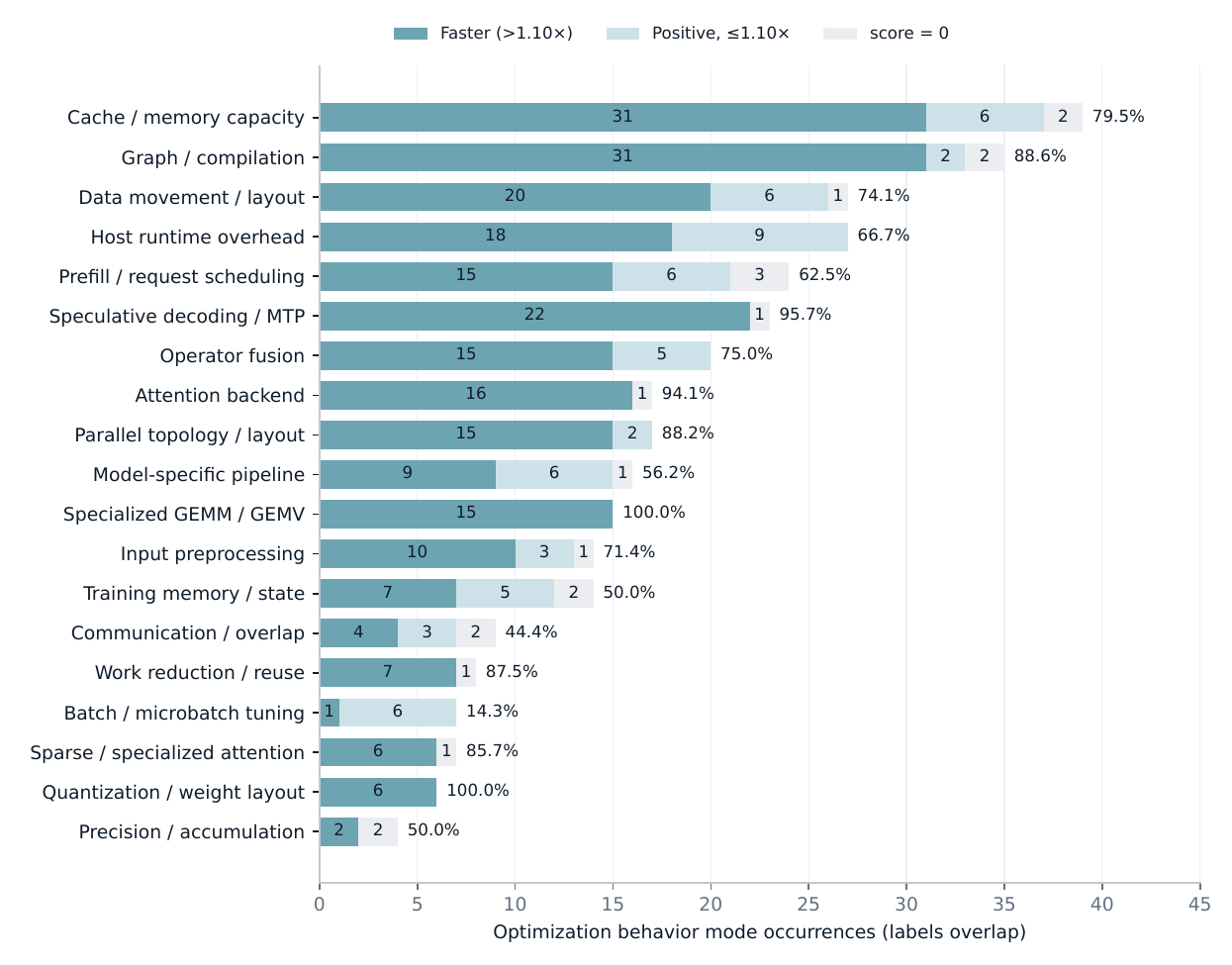}
    \caption{\textbf{Outcomes of optimization behavior modes.} Multi-label
    counts over 130 formally scored submissions, ordered by total frequency: Claude contributes 10 tasks
    and each other model contributes 20. Labels require a change present
    and enabled in the final OCI artifact; baseline features, reverted
    experiments, and disabled code are excluded. Right-hand percentages
    report the Faster fraction within each category. Sequence-state path is omitted.}
    \label{fig:successful-methods}
    \label{fig:behavior-method-outcomes}
\end{figure}
\FloatBarrier

\paragraph{Successful cases.}
\revision{Figure~\ref{fig:successful-methods} relates submitted changes to
validated outcomes. Graph/compilation changes occur in 31 Faster
submissions, cache/memory changes in 31, and speculative decoding/MTP in
22. Among all submissions with each label, the corresponding Faster
fractions are 88.6\% (31/35), 79.5\% (31/39), and 95.7\% (22/23).
These results highlight opportunities to improve execution and memory
reuse within existing engines. Kernel work is also prominent: all 15
submissions with specialized GEMM/GEMV changes are Faster, and operator
fusion accompanies 15 Faster submissions out of 20 occurrences. These
findings suggest that different optimization paths vary in their implementation
difficulty and performance gains. Effectively identifying optimization
opportunities within inference and training engines and selecting appropriate methods for specific bottlenecks could therefore enhance agents' end-to-end optimization capabilities.}

\revision{Concrete examples span multiple layers. Astra's GLM-5.3
submission introduces a small-head sparse-attention kernel and achieves
$1.7958\times$ validated speedup. Qwen's Inkling submission combines
native MTP with a larger GPU KV pool and hierarchical host caching,
achieving $3.3276\times$. Claude's LLaDA submission combines block-causal
prefill with vectorized decoding and state reuse, reaching
$6.1062\times$. }

\paragraph{Failure cases.}
\revision{The 44 non-Faster results comprise 35 positive scores at or below
$1.10\times$ and nine formal zeros.
The nine zeros consist of four unsuccessful deployments, four evaluations
without a passing quality gate, and one compliance rejection. For example,
Qwen's SFT submission records a raw $1.2540\times$ throughput ratio but
receives zero because its quality gate does not pass. Seed's TTS submission
passes the speech-quality gate but is rejected after changing the repetition
penalty. These cases show why performance, quality, and compliance are
assessed separately. Appendix~\ref{app:optimization-behavior} gives the
classification criteria and outcome decomposition.}

\FloatBarrier
\begin{figure}[!htbp]
    \centering
    \includegraphics[width=\linewidth]{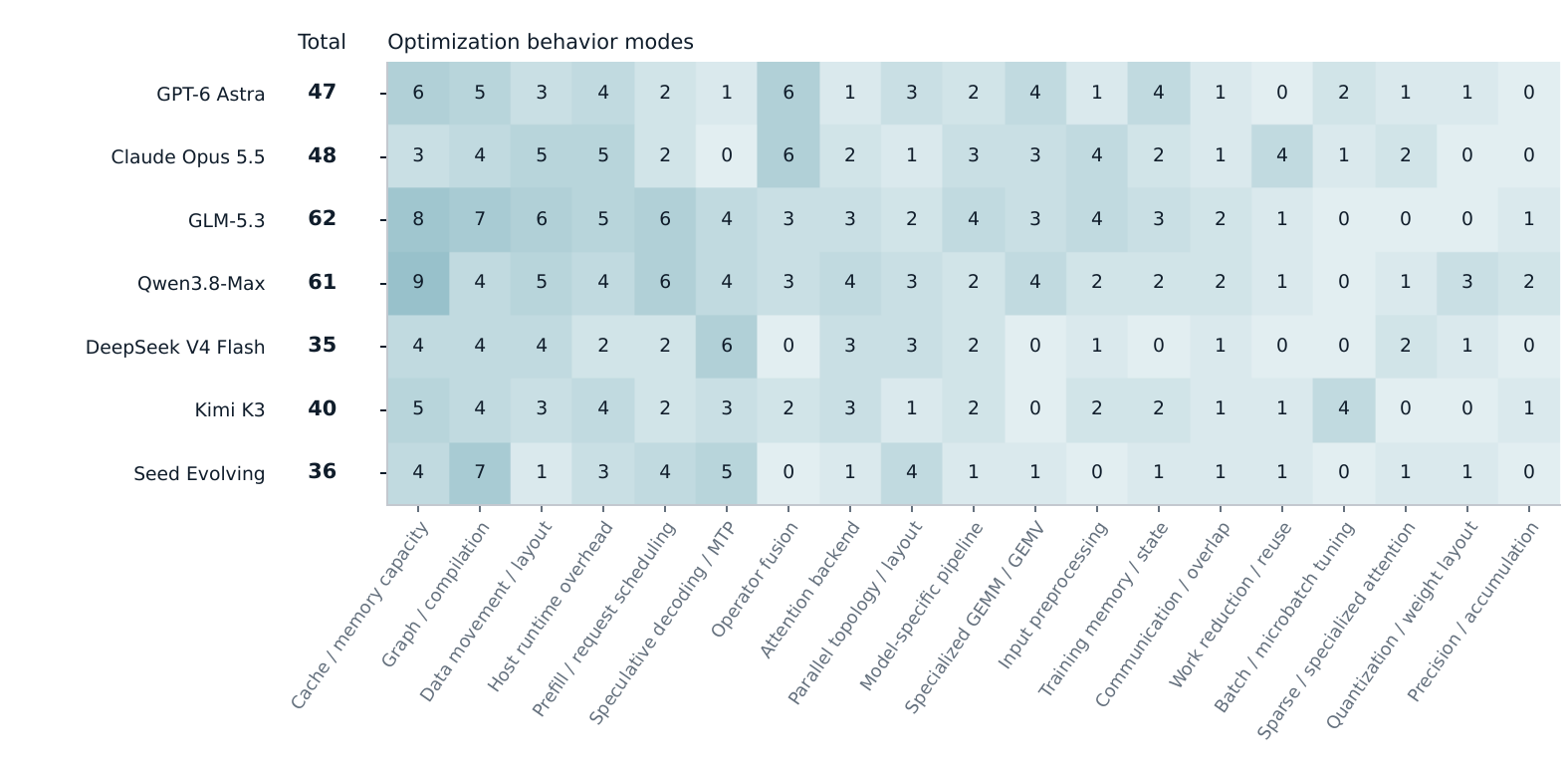}
    \caption{\textbf{Optimization behavior modes by model.}
    All 19 categories displayed in Figure~\ref{fig:successful-methods},
    ordered by total frequency. Cell values count submissions; labels overlap.
    Total is the row sum across all 19 categories, counting occurrences across tasks.
    Claude has 10 scored tasks; the other models have 20.}
    \label{fig:behavior-model-summary}
\end{figure}
\FloatBarrier

\paragraph{\revision{Optimization preferences and efficiency.}}
\revision{Figure~\ref{fig:behavior-model-summary} reveals different
optimization behavior preferences across models. Astra and Claude place
a stronger emphasis on operator fusion: each applies it on six evaluated
tasks, compared with three each for GLM and Qwen. In contrast, cache/memory
optimization is the most frequent mode for GLM and Qwen.}

\revision{GLM and Qwen exhibit the largest aggregate numbers of optimization
behavior mode occurrences, 62 and 61, and spend an average of 5\,h 26\,min
and 6\,h 53\,min per task, respectively (Table~\ref{tab:main-results}).
Astra and Claude record smaller totals of 47 and 48 across their respective
evaluated sets, yet reach an 80\% Faster Rate in only 3\,h 36\,min and
3\,h 25\,min per task.
The combination of high scores and short execution times demonstrates
strong capability to identify and implement effective optimizations.
Both models use less than half of the available eight-hour budget,
suggesting an opportunity to further improve performance through methods
that sustain profiling, experimentation, and refinement after an initial
successful solution. Encouraging this sustained optimization may unlock
a higher performance ceiling than the present runs realize.}

\FloatBarrier
\begin{figure}[!htbp]
    \centering
    \includegraphics[width=\linewidth]{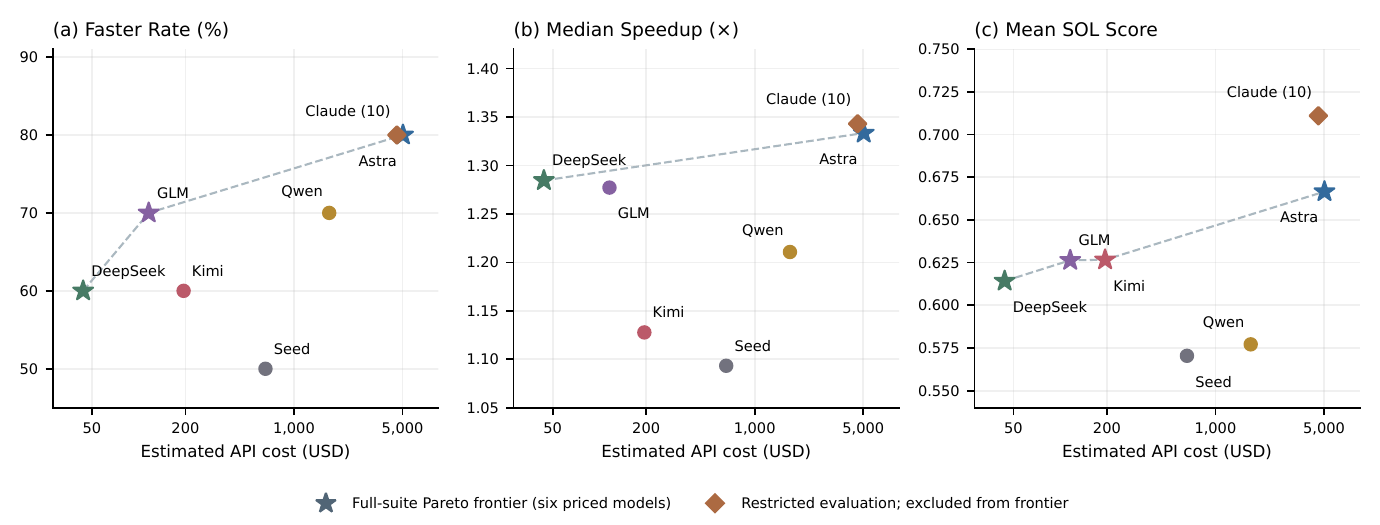}
    \caption{\textbf{Cost--efficiency trade-offs.} Estimated API token cost
    versus Faster Rate, Median Speedup, and Mean SOL Score. Stars and dashed
    lines identify the Pareto frontier among the six priced complete-suite
    configurations.}
    \label{fig:cost-performance-pareto}
\end{figure}
\FloatBarrier

\paragraph{Cost--performance trade-offs.}
\revision{Figure~\ref{fig:cost-performance-pareto} compares API cost and
optimization performance.\footnote{\revision{Costs use OpenRouter catalog prices
retrieved on November 13, 2026 (\url{https://openrouter.ai/api/v1/models}),
except Seed, which uses its official catalog price
(\url{https://docs.volcengine.com/docs/ark/model-pricing}).}}
All input is charged at the ordinary
prompt rate without cache discounts, and request-level long-context price
tiers are applied where specified.}

\revision{Among the six priced full-suite configurations, DeepSeek, GLM,
and Astra form the Faster Rate frontier. DeepSeek and Astra form the
Median Speedup frontier. The mean-SOL frontier also includes Kimi, whose
unrounded score (0.62667) is slightly higher than GLM's (0.62639).
Astra provides the strongest full-suite performance across all three
metrics, while DeepSeek and GLM offer substantially lower estimated API
costs. The frontier therefore depends on the performance metric and the
stated price schedule.}
\FloatBarrier
\paragraph{Scope of the experiment.}
The comparison reports one run per eligible model--task pair. It does not
provide repeated-run confidence intervals or a controlled comparison of
agent budgets, so the rankings should be interpreted as point estimates for
this benchmark suite.

\section{\revision{Analytical Performance References and the Lite Set}}
\label{sec:lite-set}

\revision{We provide analytical speed-of-light (SOL) references for all
20 tasks on their assigned hardware, alongside a five-task Lite Set on a
single H20: Qwen3.8-27B, Muse-Glimmer-30B, Qwen3-VL-32B,
LLaDA2.1-mini-16B, and Wan2.2-TI2V-5B. The Lite Set is designed for
evaluation under a limited compute budget of one GPU per task while
maintaining broad coverage of model architectures and execution engines.
It covers text,
vision--language, diffusion-language, and video workloads; its measured
agent speedups are drawn from the corresponding full-suite tasks.}

\paragraph{Analytical SOL reference.}
\revision{We use AISimulate~\citep{aisimulate} to estimate operator costs
from the assigned H20 or Ascend 910B2 hardware specifications, including
compute throughput, memory bandwidth, and the deployed precision.
Native execution records supply scheduling, cache behavior, workload
shapes, and routing where available; fixed computation graphs supply
video-generation and other model-specific stages. Training models include
rollout, forward/backward computation, updates, and synchronization as
applicable. Table~\ref{tab:sol-references} reports baseline throughput $b$,
analytical reference $s$, and their ratio $b/s$ in each task's native units.
This expands reference-normalized evaluation to both inference and training.}

\begin{table}[!htbp]
\centering\small
\caption{Baseline throughput $b$ and analytical SOL reference $s$ for all 20 tasks on their assigned hardware. $b/s$ is the baseline fraction of the analytical reference. $^{*}$marks the five single-H20 Lite tasks; $^{\dagger}$marks prefix-average extrapolations. Units apply to both $b$ and $s$; training rates count complete workload steps.}
\label{tab:lite-set}\label{tab:sol-references}
\begingroup\fontsize{8}{10}\selectfont\setlength{\tabcolsep}{4pt}
\begin{tabular*}{\linewidth}{@{\extracolsep{\fill}}lllrrr@{}}
\toprule
Task & Hardware & Unit & Baseline $b$ & SOL $s$ & $b/s$ \\\midrule
DeepSeek V4 Flash & $8\times$910B2 & tokens/s & 160.6205 & 1083.4008 & 0.1483 \\
GLM-5.2$^{\dagger}$ & $32\times$910B2 & tokens/s & 12.9325 & 658.0390 & 0.0197 \\
GLM-5.3 FP8 & $16\times$H20 & tokens/s & 113.4608 & 378.2019 & 0.3000 \\
GLM-5.3 Flash$^{\dagger}$ & $16\times$910B2 & tokens/s & 4.8850 & 176.8834 & 0.0276 \\
Inkling 975B & $8\times$H20 & tokens/s & 104.3700 & 586.6227 & 0.1779 \\
Kimi K3$^{\dagger}$ & $32\times$H20 & tokens/s & 26.2467 & 201.4855 & 0.1303 \\
LingBot World 2.0 & $8\times$H20 & frames/s & 7.5420 & 10.7843 & 0.6993 \\
LLaDA2.1-mini$^{*}$ & $1\times$H20 & tokens/s & 92.7092 & 642.2889 & 0.1443 \\
MiniMax-H3 & $2\times$H20 & videos/h & 202.7969 & 301.5330 & 0.6726 \\
Muse-Glimmer 30B$^{*}$ & $1\times$H20 & tokens/s & 47.8075 & 71.1419 & 0.6720 \\
Nemotron 3 Ultra$^{\dagger}$ & $8\times$H20 & tokens/s & 26.6906 & 136.2664 & 0.1959 \\
$\pi_{0.5}$-LIBERO & $1\times$H20 & chunks/s & 2.7078 & 31.2171 & 0.0867 \\
Qwen3.8-27B$^{*}$ & $1\times$H20 & tokens/s & 108.9275 & 300.2242 & 0.3628 \\
Qwen3-TTS Base & $1\times$H20 & ref. audio s/s & 9.8315 & 47.5556 & 0.2067 \\
Qwen3-VL-32B$^{*}$ & $1\times$H20 & tokens/s & 106.0794 & 301.4007 & 0.3520 \\
Wan2.2-TI2V-5B$^{*}$ & $1\times$H20 & videos/h & 6.6280 & 10.2098 & 0.6492 \\
\midrule
Qwen3-30B-A3B GRPO & $8\times$H20 & steps/h & 5.8887 & 16.1602 & 0.3644 \\
MiMo-7B-RL SFT & $8\times$H20 & steps/h & 1472.2561 & 2228.1136 & 0.6608 \\
Qwen3-8B OPD & $16\times$H20 & steps/h & 53.7653 & 78.7682 & 0.6826 \\
Qwen3-VL-8B GRPO & $8\times$H20 & steps/h & 4.5090 & 7.3751 & 0.6114 \\
\bottomrule\end{tabular*}
\endgroup
\end{table}
\FloatBarrier

\paragraph{Reference normalization.}
\begingroup
Following the runtime-score form of SOL-ExecBench~\citep{lin2026solexecbench},
let $t_b$, $t$, and $t_{\mathrm{sol}}$ be the times required by the baseline,
validated agent submission, and analytical reference for a common nominal
work amount $W$. Thus $t_b=W/b$, $t=W/p$, and $t_{\mathrm{sol}}=W/s$,
where $p$ is candidate throughput. We use
\begin{equation}
 S(t)=\frac{t_b-t_{\mathrm{sol}}}
 {(t-t_{\mathrm{sol}})+(t_b-t_{\mathrm{sol}})},
 \qquad
 \operatorname{SOLScore}=\begin{cases}
 0, & \text{formal zero-credit outcome},\\
 1, & \text{validated outcome with }t\leq t_{\mathrm{sol}},\\
 S(t), & \text{validated outcome with }t>t_{\mathrm{sol}}.
 \end{cases}
 \label{eq:sol-score}
\end{equation}
The baseline maps to 0.5 and the SOL reference maps to 1. All task
references satisfy $s>b$, and the piecewise rule bounds the score in
$[0,1]$. Scores are capped before evaluating the fractional expression
when a submission reaches or exceeds the reference.
\endgroup

\paragraph{Model results.}
\revision{We compute the arithmetic mean of per-task SOL scores, including
formal zeros. Table~\ref{tab:main-results} reports full-suite and Lite means
with their coverage: six models use 20 and five tasks, respectively;
Claude uses 10 and three. Astra has the highest complete-suite mean
(0.667), while Qwen has the highest complete Lite mean (0.670).
Capping makes the aggregate bounded while preserving the
original speedup in the underlying result record.}

\section{Conclusion}
\label{sec:conclusion}
We introduced \ours, a benchmark of 20 production-oriented training and
inference tasks with stack-wide optimization, OCI-layer delivery, and
independent quality and performance evaluation. \revision{Seven agent--model
configurations evaluated with 1M-token context limits and an eight-hour
budget yield 86 Faster results among 130 formally scored entries.
GPT-6 Astra leads the complete-suite comparison with an 80\% Faster Rate
and a 0.667 mean SOL score. Analytical
references for all 20 tasks and the single-H20 Lite Set support evaluation
across both production-scale and accessible workloads.} \ours provides a testbed for advancing this infrastructure component of recursive
self-improvement.

\bibliographystyle{iclr2027_conference}
\bibliography{main}

\clearpage
\beginappendix
\section{Task Specifications and Baseline Recipes}
\label{app:tasks}

This section of appendix specifies input construction, training objectives, and measurement
boundaries. Model and hardware assignments and workload cardinalities are given
in Table~\ref{tab:task-overview}; acceptance thresholds are given in
Appendix~\ref{app:quality-gates}.

\subsection{Reference Recipes and Optimization Boundaries}
\label{app:task-common}

Each task starts from an executable deployment or training recipe with its
prescribed checkpoint, tokenizer or processor, numerical representation, and
resource allocation. These define a reference operating point, not an optimal
implementation. Inference recipes retain the supplied weight representation,
including low-precision checkpoints where specified. Training recipes retain the
initial model state, objective, effective batch, masks, optimizer hyperparameters,
and update schedule. Engine scheduling, microbatching, kernel fusion, memory
layout, and communication overlap may be optimized within these constraints.
Changing a reduction order is permitted; changing the objective or omitting
required computation is not. Quality evaluation and artifact validation follow
Appendix~\ref{app:protocol}.

\subsection{Inference Task Inputs and Generation Settings}
\label{app:inference-tasks}

\paragraph{Text replay.}
The serving workload is constructed from recorded Terminal-Bench agent sessions.
Each request retains its input length, output budget, relative arrival time, and
prefix-block identifiers. Sessions are selected to fit the task's context limit.
Recorded timestamps determine request order but do not impose arrival-time
delays during ASAP replay. A deterministic generator reconstructs token sequences in
128-token blocks, reusing a block for the same prefix identifier. Thus, replay
preserves sequence lengths and a controlled prefix-sharing pattern, rather than
the original conversation text. It tests serving efficiency; answer quality is
assessed separately.

Requests are dispatched as soon as concurrency capacity is available (ASAP),
without waiting for their recorded arrival times, subject to a maximum of four
concurrent requests for full-trace replay and one for the windowed workloads.
The random seed is 42. The evaluator supplies token IDs
directly, checks their lengths against service-reported input counts, and uses
service-reported output-token counts instead of re-tokenizing generated strings.
Generation uses near-zero temperature and the recorded output budget. Full-trace
measurement includes queueing and draining outstanding
requests. Windowed measurement instead counts tokens received before the cutoff,
including verified cumulative counts from a request still in flight; unserved
requests beyond the cutoff are not treated as completed requests. These are two
different measurement contracts and use separate reference denominators.

\paragraph{Vision--language requests.}
For throughput, each request combines one deterministically generated image with
synthetic text of nominal length 2048 tokens and an output budget of 256 tokens;
the driver uses concurrency eight and seed 42. Image processing and multimodal
prefill remain in the serving path. These synthetic performance inputs are
separate from the natural image--question pairs used for quality evaluation.

\paragraph{Video and interactive generation.}
MiniMax-H3 uses text-to-video-and-audio generation of a golden retriever running
through a meadow, with seed 42, video flow shift 12, and audio flow shift 3.
Wan uses text-to-video generation with the original, non-distilled weights: two
prompts (boxing cats and a dog in a meadow), each paired with seeds 42 and 43.
Its UniPC solver uses guidance scale 5 and flow shift 5, with a fixed negative
prompt suppressing blur, static scenes, distortion, and artifacts. Both tasks
produce media at 24 frames/s and retain the generation lengths in
Table~\ref{tab:task-overview}.

LingBot starts from a fixed lakeside image, prompt, and seed 42. The driver
changes camera controls after chunk indices 3 and 12, changes the prompt to a
snowy scene after index 6, and jointly changes the prompt and controls to a
sunset scene after index 9 (indices start at zero). Events are injected at chunk
boundaries, so their effect is independent of wall-clock speed. Guidance is 1;
upscaling and frame interpolation are disabled. The entire native frame stream,
including the initial chunk, contributes to performance.

\paragraph{Actions and speech.}
Action requests carry the observation images, robot state, language instruction,
and explicit denoising noise. The evaluator selects observations from the
reference pool and reuses the same inputs across timed rounds. Requests are
serial, matching batch-one policy execution. Speech requests carry target text,
reference audio, and its transcript for voice cloning, with seed 42, unit speech
speed, and a maximum of 2048 generated audio tokens. The service returns streaming
PCM. Throughput uses the \emph{fixed reference audio duration} as its numerator,
so lengthening the candidate waveform cannot improve the score.

\subsection{Training Task Data and Objectives}
\label{app:training-tasks}

\paragraph{Training tasks.}
The four training workloads use pinned data and objective contracts:
Qwen3-30B-A3B-Base uses the verl DAPO-style
\emph{DAPO-Math-17K} recipe on one node with eight H20 GPUs, Megatron training,
and vLLM rollout through SkyRL-Tinker, including the prescribed chat template,
response limits, group sampling, math reward, advantage normalization,
old-policy pass, and dual-clipped objective; MiMo-7B-RL uses the pinned verl
Megatron \emph{MTP-SFT} recipe through SkyRL-Tinker on eight H20 GPUs, with the
GSM8K main split, native chat template, masked assistant targets, MTP auxiliary
loss, sampler, optimizer schedule, and warmup boundary, without rollout or a
teacher service; the Qwen3-8B student uses the original verl \emph{on-policy
distillation (OPD)} recipe with FSDP training and vLLM rollout, while a frozen
Qwen3-32B teacher scores sampled tokens in a separate vLLM pool across two
nodes, using the pinned GSM8K and MATH prompts, sampling limits, round-robin
order, and importance-weighted teacher--student objective; and Qwen3-VL-8B-
Instruct uses the official verl \emph{Geo3K} recipe through verl-tinker with an
FSDP2 actor, frozen reference model, and colocated vLLM rollout on eight H20
GPUs, retaining the image-preserving prompt format, response limits, mathruler
reward, grouped sampling, dual-clipped policy loss, reference KL term, and
multimodal forward passes. Tokenization, filtering, input-hash checks, and
recipe-defined reductions occur as specified by the benchmark implementation;
the VL test split is reserved for quality evaluation.

\paragraph{Workload steps and performance baselines.}
A policy-training step includes weight synchronization, fresh rollout,
reward or teacher scoring, input construction, required old-policy/reference
forward passes, and every minibatch update. All optimizer futures must finish
before the step is complete. SDK request splitting and microbatching cannot
change the logical loss denominator or update count. Sampling uses temperature
one, top-$p=1$, and the task-defined seed-42 schedule. Fixed filtering rules
and manifests account for responses, usable datums, tokens, and updates;
candidates cannot improve throughput by reducing prescribed work.

The current baseline rates are 5.888738, 1472.256060, 53.765306, and
4.509013 workload steps/hour for text GRPO, MTP SFT, OPD, and VL GRPO,
respectively. Text GRPO uses the median of three independent cold-start
measurements; MTP SFT and VL GRPO use five. OPD retains its single-sample
calibration. Short Oracle smoke runs do not replace these denominators. Online rollout lengths
vary, and short RL windows need not estimate long-training steady-state throughput.
Quality replay uses frozen inputs, whereas performance uses fresh rollouts.
Periodic validation, checkpointing, and long-training epochs from upstream
recipes are outside these explicitly defined short workloads.

\subsection{Measurement Boundaries}
\label{app:task-timing}

Table~\ref{tab:task-timing} specifies the work included in each denominator.
Service startup, model loading, and the stated warmup are outside measurement.
Warmup is executed once per evaluation, not subtracted from individual request
latencies. Client-observed intervals include communication with the evaluated
service. Reference and candidate use the same boundaries; internal kernel or
forward-pass timings are diagnostic only.

\begin{table}[htbp]
\centering
\small
\caption{Warmup and end-to-end timing contracts. Quality scoring is outside the
performance interval.}
\label{tab:task-timing}
\begingroup
\setlength{\tabcolsep}{4pt}
\renewcommand{\arraystretch}{1.16}
\begin{tabular}{@{}>{\raggedright\arraybackslash}p{0.19\linewidth}
                  >{\raggedright\arraybackslash}p{0.29\linewidth}
                  >{\raggedright\arraybackslash}p{\dimexpr0.52\linewidth-16pt\relax}@{}}
\toprule
\rowcolor{gray!12}
\textbf{Workload} & \textbf{Warmup} & \textbf{Measured interval}\\
\midrule
Text replay & One request for full replay; none for GLM-5.2; one short request for GLM-5.3-Flash.
& Full replay: start of ASAP dispatch to final completion. Windowed replay: ASAP dispatch to cutoff (or exhaustion where allowed), using only tokens received in that interval.\\
\addlinespace[3pt]
Vision--language & No separate request.
& Dispatch and completion of the complete multimodal request batch.\\
\addlinespace[3pt]
Video generation & One two-denoising-step request for MiniMax; one full-schedule request with a separate seed for Wan.
& Synchronous generation requests through receipt of media files, including transfer and file writing; decoding and perceptual comparison follow timing.\\
\addlinespace[3pt]
Interactive world & A separate two-chunk session.
& Initialization message to completion of the final native frame chunk; event injection is included.\\
\addlinespace[3pt]
Action generation & Two requests.
& Serial request loop over all timed rounds, including receipt and parsing of action chunks.\\
\addlinespace[3pt]
Speech synthesis & One request.
& First timed request to receipt of the last PCM stream. Request preparation, audio persistence, and ASR are outside timing.\\
\addlinespace[3pt]
Training & Complete workload steps.
& All timed workload iterations through completion of their optimizer futures; includes rollout weight synchronization and teacher scoring where applicable. SFT data preparation and quality replay are outside timing.\\
\bottomrule
\end{tabular}
\endgroup
\end{table}

Warmup updates in training are retained: timed steps continue from the resulting
model and optimizer state. Checkpoint recovery probes and curve comparisons are
not inserted into the timed run. Execution records retain completed requests,
rollouts, usable training examples, tokens, and optimizer updates so that a
reduction in the amount of required work cannot be mistaken for a systems gain.

\section{Detailed Evaluation Protocol}
\label{app:protocol}

\subsection{Task-Specific Quality Gates}
\label{app:quality-gates}
Performance gains are accepted only when the submission satisfies the task's
quality requirements. An independent driver evaluates the submitted system
against fixed task inputs and reference assets. The gate depends on the
workload: answer quality for language understanding, reference-output
consistency for generated media and actions, speech intelligibility for
synthesis, and numerical training behavior for training tasks.

\paragraph{Language and vision-language understanding.}
Eight pure-text inference tasks use LongBench v2\footnote{\url{https://github.com/THUDM/LongBench}}:
DeepSeek-V4-Flash, GLM-5.3, Inkling, Kimi K3, LLaDA2.1-mini,
Muse-Glimmer, Nemotron-3-Ultra, and Qwen3.8. GLM-5.2 and GLM-5.3-Flash
instead use GSM8K, with accuracy lower bounds of 0.88 and 0.865,
respectively. Their long-context performance traces do not determine the
choice of quality benchmark.
The candidate and baseline are evaluated under the same task-specific
data selection, prompt, context-length policy, generation settings, and
answer-scoring procedure. The gate checks accuracy against the task-specific
acceptance threshold, separately from throughput measurement.
LLaDA2.1-mini uses a 36-question subset at the 24k gate tier;
DeepSeek-V4-Flash and Muse-Glimmer use 100-question subsets at 128k,
and GLM-5.3, Inkling, Kimi K3, Nemotron-3-Ultra, and Qwen3.8 use
100-question subsets at 256k.
These gate tiers are separate from the performance traces in
Table~\ref{tab:task-overview}.
For Qwen3-VL-32B, we evaluate image--question pairs on the 847 multiple-choice
validation examples of MMMU~\citep{yue2024mmmu}, extract the predicted option,
and compute answer accuracy. The current gate accepts an absolute deviation
of at most 0.05 from the calibrated reference accuracy of 0.7273.

\paragraph{Video generation.}
Wan2.2-TI2V and MiniMax-H3 use fixed prompts, seeds, and generation settings.
We decode the returned media and check its dimensions, frame count, frame
rate, and encoding before comparing it with reference outputs using
LPIPS~\citep{zhang2018perceptual}.
Wan2.2 uses four prompt--seed pairs, generating videos at
1280$\times$704 resolution, 121 frames, and 24 frames/s for comparison with
the corresponding reference videos. Each video must satisfy all calibrated
bounds in Table~\ref{tab:wan-quality}, rather than only on average across
cases. Mean LPIPS averages the framewise distances; worst-frame LPIPS takes
their maximum. For temporal consistency, the verifier computes the difference
between consecutive frames in each video, then compares these temporal
differences against the reference using the mean absolute error for each
adjacent-frame pair. The gate checks both the mean and the maximum of these
per-pair errors. The motion ratio is the mean absolute temporal difference
of the candidate divided by that of the reference. Pixel values for these
temporal metrics are normalized to $[0,1]$. Frame timestamps and duration
are checked as well.

\begin{table}[htbp]
    \centering
    \small
    \caption{Wan2.2 quality bounds. Decimal values are rounded for display;
    the task configuration retains the full-precision calibrated values.}
    \label{tab:wan-quality}
    \begin{tabular}{ll}
        \toprule
        Measurement & Acceptance bound \\
        \midrule
        Mean framewise LPIPS & $\leq 0.25834550$ \\
        Worst-frame LPIPS & $\leq 0.31249873$ \\
        Mean temporal-difference absolute error & $\leq 0.05224669$ \\
        Worst-pair temporal-difference mean absolute error & $\leq 0.14031570$ \\
        Motion magnitude ratio & $[0.97985323,\,1.04997272]$ \\
        Absolute duration error & $\leq 0.001$\,s \\
        Absolute frame-timestamp error & $\leq 0.0001$\,s \\
        \bottomrule
    \end{tabular}
\end{table}

MiniMax-H3 additionally checks the generated audio: mean video LPIPS must
be at most 0.2, audio spectral cosine similarity at least 0.8, and the audio
root-mean-square amplitude ratio between 0.5 and 2.0. Its audio format must
match the prescribed sample rate and channel count. All required checks
must pass.

\paragraph{Interactive world modeling.}
For LingBot-World 2.0, the driver replays a fixed initial image, prompt, seed,
and sequence of camera and prompt-update events over 16 chunks totaling
189 frames. Each chunk must have the expected frame count, shape, and
effective event identifier. The first, middle, and last frames of each chunk
are compared with the
reference session: mean absolute pixel error must be at most 2.0 on the
8-bit scale and peak signal-to-noise ratio at least 35\,dB in every chunk.
Missing or additional chunks fail the gate.

\paragraph{Action generation.}
For $\pi_{0.5}$-LIBERO, we randomly select 16 observations from a frozen pool
of 64 reference cases and provide the same observation, instruction, and
denoising noise to the candidate. Every returned action chunk must have the
reference shape, maximum absolute error at most 0.05, and mean absolute
error at most 0.01. This gate measures action-output consistency with the
baseline; closed-loop robot success is not evaluated.

\paragraph{Speech synthesis.}
Qwen3-TTS-Base is evaluated on 202 fixed Chinese Seed-TTS cases with reference
audio and transcripts for voice cloning. Every request must complete and
return valid, nonempty audio. A fixed Paraformer-zh recognizer transcribes
the generated speech, and the official Seed-TTS normalization and scoring
procedure computes word error rate (WER). The arithmetic mean of per-utterance
WER must be at most 1.4\%. Speech recognition and audio persistence are
outside the performance timing interval. This gate measures intelligibility;
speaker similarity and subjective naturalness are not acceptance criteria.

\paragraph{Training replay and admission.}
Each gate starts from the pinned model initialization in a fresh Train Job,
separate from the performance job. Frozen inputs, reference signals, and
calibration assets are bound to the model, objective, optimizer schedule,
workload identity, and file hashes. Invalid identities fail admission before
scoring. The trusted CPU driver reconstructs losses from returned forward
outputs and records synchronized global gradient norms before clipping.
Missing, duplicated, reordered, or non-finite steps are rejected, as are
changes to masks, input counts, or optimizer-update counts.

Text GRPO replays two frozen rollout batches of 4096 datums, giving 64
continuous optimizer updates (\texttt{rl-curve/5}). MiMo SFT replays 16
batches of 64 examples, covering updates beyond LR warmup
(\texttt{train-curve/1}); the MTP contribution remains active in training
and in the global gradient norm. OPD replays eight batches of 128 datums
(\texttt{train-curve/1}), with frozen sampling log-probabilities and teacher
advantages, without making fresh teacher or rollout calls. These three
gates require relative loss error strictly below 2\% and relative
gradient-norm error strictly below 15\% at every update. Relative error is
$|x-x_{\mathrm{ref}}|/|x_{\mathrm{ref}}|$; a zero reference requires a zero
candidate value.

\paragraph{Vision--language training gate.}
VL GRPO uses \texttt{rl-curve/6}: two frozen batches of 2560 datums,
eight consecutive 640-datum updates, and seven required signal families.
It checks total loss, global gradient norm, policy loss, KL loss, actor
forward fidelity, reference forward fidelity, and a final isolated KL
gradient probe. The probe uses the final canonical minibatch with zero
advantages and a fixed reference-logprob offset of one. Learning rate zero
allows measurement of the native KL gradient norm without changing the
model; the probe is outside performance timing.

For the five reference-comparison families, the maximum pointwise error is
$e=\max_k |x_k-r_k|/[a+b\max(|r_k|,c)]$.
Table~\ref{tab:vl-gate} lists the calibrated constants. Actor and reference
forward signals are already nonnegative errors and are checked directly
against upper bounds, rather than penalizing improved agreement. All
families must pass; values between the pass and fail boundaries are
inconclusive and receive no validated performance credit. The published
policy does not automatically retry inconclusive gates.
\begin{table}[htbp]
\centering
\small
\caption{VL GRPO gate constants. Pass means $e\leq P$; fail means
$e\geq F$. For forward checks, $e$ is the driver's maximum absolute
log-probability discrepancy.}
\label{tab:vl-gate}
\begin{tabular}{lrrrrr}
\toprule
Signal family & $a$ & $b$ & $c$ & $P$ & $F$ \\
\midrule
Total loss & 0.002 & 0.02 & 0.01 & 0.5 & 1 \\
Gradient norm & 4 & 0.10 & 0.1 & 0.5 & 1 \\
Policy loss & 0.002 & 0.02 & 0.01 & 0.5 & 1 \\
KL loss & 0.0003 & 0.20 & 0 & 0.5 & 1 \\
KL gradient probe & 0.002 & 0.05 & 0 & 0.5 & 1 \\
Actor forward & --- & --- & --- & 6 & 8 \\
Reference forward & --- & --- & --- & 0.6 & 0.9 \\
\bottomrule
\end{tabular}
\end{table}

\paragraph{Calibration controls and scope.}
Training admission includes independent reference replay and a benign
known-good implementation, together with no-update and wrong-learning-rate
negative controls. MiMo additionally tests missing MTP; VL GRPO tests
missing KL. These controls validate the selected gate for the pinned
workload, not arbitrary changes to training semantics. The gates test local
numerical behavior rather than downstream convergence or generalization.
All four tasks install hash-pinned baseline support files before applying
the candidate OCI layer. Candidate modifications and deletions remain
effective; trusted CPU evaluation code and reference assets remain outside
the candidate environment.

\paragraph{Independent review for potential benchmark hacking.}
In addition to the executable quality checks, an independent reviewer examines
the submitted changes and their accompanying summary. The review
assesses whether the changes constitute legitimate infrastructure optimizations
that transfer to general workloads, and flags non-generalizable shortcuts or
quality-sacrificing changes that violate the task contract. Limited or absent
speedup alone does not make a submission non-compliant.

\section{Agent Configurations and Reproducibility}
\label{app:agents}

\paragraph{Configurations and execution budget.}
We evaluate seven model configurations: GPT-6 Astra, Claude Opus~5.5,
DeepSeek V4 Flash, GLM-5.3, Qwen3.8-Max, Seed Evolving, and Kimi K3.
\revision{Each run receives an eight-hour execution budget, English task
instructions, \texttt{reasoning=max}, and a 1M-token agent context limit.
GPT-6 Astra runs in Codex, Claude Opus~5.5 in Claude Code, and the remaining
models in OpenCode. Claude has formal scores on 10 of the 20 tasks.
The other tasks encounter the following API error:}
\begin{quote}\revision{\small API Error: 400 This request may violate the
Claude model's safety policy and has been terminated early.}\end{quote}
\revision{The excluded tasks are DeepSeek V4 Flash, GLM-5.2, GLM-5.3 FP8,
GLM-5.3 Flash, Kimi K3, Muse-Glimmer, and Nemotron inference;
Qwen3.8 inference; Qwen3-8B OPD; and Qwen3-VL-8B GRPO.
The 10 formally scored tasks determine all Claude aggregates, including
cost and mean execution time. Its Lite mean uses LLaDA2.1-mini,
Qwen3-VL, and Wan. Formal zero is a scored outcome and is never removed
from aggregation.}

\paragraph{Submission and evaluation.}
Each run receives a task instruction, resource allocation, baseline image,
model and data assets, verifier configuration, and performance baseline.
Agents submit \texttt{submit\_oci.tar.gz}, containing an executable
\texttt{launch.sh} and the modifications allowed by the task. The layer is
validated and applied to a clean baseline with the same contents on every
rank. For training, pinned support files are installed before the candidate
layer; candidate modifications remain effective while trusted evaluation
code and reference assets remain outside the candidate environment.

\paragraph{Inputs and timing.}
Task-defined seeds and frozen inputs are controlled by the evaluator. The
shared inference drivers use seed 42; training uses the prescribed training
and sampling seeds. Quality replay and performance are distinct: training
replay uses fixed inputs, while performance includes fresh rollout and all
required update work. The same workload and timing boundaries apply to the
baseline and candidate. Startup, model loading, and quality evaluation are
outside the performance denominator.

\paragraph{Aggregation and reporting.}
One run supplies each eligible model--task result; the comparison does not
select the best score from repeated attempts. Valid optimizations retain their
raw throughput ratios, including slowdowns, while evaluated failures receive
zero credit. Faster Rate uses the strict $1.10\times$ threshold, and Median
Speedup uses the same eligible set, as specified in Section~\ref{sec:scoring}.
Single-run results are point estimates for this benchmark suite.

\revision{Token totals include all recorded input and output, with reasoning
counted once according to the harness's usage schema. Cached and repeatedly
processed input is included. Mean duration uses the agent-execution interval,
excluding environment setup and independent verification. For verification-only
records, the interval is taken from the originating agent run. API costs use
the OpenRouter catalog snapshot and per-request tiers described in
Figure~\ref{fig:cost-performance-pareto}, with the official catalog price
used for Seed. Device allocation and
verifier overhead are excluded from token, time, and API-cost accounting.}

\section{Optimization Behavior and Outcome Analysis}
\label{app:optimization-behavior}

\paragraph{Population and evidence.}
\revision{We review all 130 formally scored model--task submissions using
their trajectories, final OCI contents, enabled launch settings, and
independent evaluation records. Claude contributes 10 tasks; the other
six models contribute 20 each. A label requires an identifiable change
relative to the supplied runnable baseline that remains enabled in the
submitted artifact. The review includes launch-parameter tuning and
source changes. Preinstalled support patches, profiling-only changes,
operational repairs, proposed or reverted changes, and code disabled by
the final configuration are excluded. In particular, the presence of a
file or a keyword alone is insufficient to establish a label.}

\paragraph{Counting rules.}
\revision{For behavior $m$, $A_m$ counts submissions with that label,
$F_m$ counts those with validated speedup strictly above $1.10\times$,
and $V_m$ counts those with a positive formal score. Each label is counted
at most once per submission; labels may overlap. Figure~\ref{fig:successful-methods}
partitions $A_m$ into Faster, positive non-Faster, and formal-zero outcomes.
Table~\ref{tab:behavior-method-outcomes} reports both $F_m/A_m$ and
$F_m/V_m$. These rates describe complete submissions and do not estimate
the success probability of trying an isolated method. Classification is
semantic review rather than an independent double-annotation study.}

\begin{table}[!htbp]\centering\footnotesize
\caption{Optimization behavior modes across 130 formally scored submissions. $A$ counts submissions with a label, $F$ counts Faster results, and $V=F+\mathrm{Low}$. Low denotes positive non-Faster scores; Zero denotes formal zero scores. Labels overlap; rates describe associations within submissions.}
\label{tab:behavior-method-outcomes}
\setlength{\tabcolsep}{4pt}
\begin{tabular}{lrrrrrr}\toprule
Behavior & $A$ & $F$ & Low & Zero & $F/A$ & $F/V$ \\\midrule
Cache / memory capacity & 39 & 31 & 6 & 2 & 79.5\% & 83.8\% \\
Graph / compilation & 35 & 31 & 2 & 2 & 88.6\% & 93.9\% \\
Speculative decoding / MTP & 23 & 22 & 0 & 1 & 95.7\% & 100.0\% \\
Data movement / layout & 27 & 20 & 6 & 1 & 74.1\% & 76.9\% \\
Host runtime overhead & 27 & 18 & 9 & 0 & 66.7\% & 66.7\% \\
Attention backend & 17 & 16 & 0 & 1 & 94.1\% & 100.0\% \\
Prefill / request scheduling & 24 & 15 & 6 & 3 & 62.5\% & 71.4\% \\
Operator fusion & 20 & 15 & 5 & 0 & 75.0\% & 75.0\% \\
Parallel topology / layout & 17 & 15 & 2 & 0 & 88.2\% & 88.2\% \\
Specialized GEMM / GEMV & 15 & 15 & 0 & 0 & 100.0\% & 100.0\% \\
Input preprocessing & 14 & 10 & 3 & 1 & 71.4\% & 76.9\% \\
Model-specific pipeline & 16 & 9 & 6 & 1 & 56.2\% & 60.0\% \\
Training memory / state & 14 & 7 & 5 & 2 & 50.0\% & 58.3\% \\
Work reduction / reuse & 8 & 7 & 0 & 1 & 87.5\% & 100.0\% \\
Sparse / specialized attention & 7 & 6 & 0 & 1 & 85.7\% & 100.0\% \\
Quantization / weight layout & 6 & 6 & 0 & 0 & 100.0\% & 100.0\% \\
Communication / overlap & 9 & 4 & 3 & 2 & 44.4\% & 57.1\% \\
Precision / accumulation & 4 & 2 & 0 & 2 & 50.0\% & 100.0\% \\
Batch / microbatch tuning & 7 & 1 & 6 & 0 & 14.3\% & 14.3\% \\
\bottomrule\end{tabular}\end{table}
\FloatBarrier

\paragraph{Coverage and dominant behaviors.}
\revision{Of 130 submissions, 104 retain at least one confirmed behavior
and 26 contain no confirmed new optimization behavior. The labeled
submissions contribute 330 incidences across 20 nonempty categories.
Graph/compilation and cache/memory changes each occur in 31 Faster
submissions, out of 35 and 39 occurrences, respectively. Speculative
decoding/MTP accompanies Faster outcomes in 22 of its 23 occurrences.
All 15 submissions labeled specialized GEMM/GEMV are Faster. These
associations show that both engine configuration and kernel work
accompany effective end-to-end optimization.}

\paragraph{Positive but insufficient gains.}
\revision{Among the 44 non-Faster submissions, 25 retain identified
behaviors: 19 have positive scores and six receive formal zeros.
The remaining 19 have no confirmed new behavior. Conversely, seven
Faster submissions have no confirmed new behavior under the artifact-based
rules. A score above the fixed reference baseline does not by itself prove
a new optimization relative to the runnable starting environment; labels
are therefore assigned independently of the score.}

\paragraph{Zero credit and interpretation.}
\revision{The nine formal zeros comprise four unsuccessful deployments
(GLM on GLM-5.3 Flash, Qwen on GLM-5.3 FP8 and Flash, and Seed on
GLM-5.3 Flash), four evaluations without a passing quality gate
(GLM text GRPO, Qwen MiMo SFT, and DeepSeek Wan and VL GRPO),
and one compliance rejection (Seed TTS). Seed's TTS submission changes
the repetition penalty and is rejected despite passing the WER gate.
The work-reduction label records that submitted behavior; it does not
endorse its compliance. Missing Claude scores are outside this population
and are never converted to formal zeros.}

\revision{Figure~\ref{fig:behavior-model-summary} uses these same 130
submissions. The heatmap displays all 19 categories shown in
Figure~\ref{fig:successful-methods}, ordered by total incidence; sequence-state
path is omitted from both figures. Heatmap cells count submissions, while
Total sums all 19 cells in each model row, counting optimization behavior
mode occurrences across tasks. Claude's 10-task coverage remains explicit.}

\paragraph{\revision{Classification criteria.}}
\revision{Table~\ref{tab:behavior-criteria} defines every nonempty category
and gives concrete evidence from submitted artifacts. The common
inclusion rule above applies to every row. Baseline features and
unsuccessful experiments that were reverted do not receive labels;
quality and compliance outcomes are recorded separately.}

\begingroup
\small
\setlength{\tabcolsep}{4pt}
\renewcommand{\arraystretch}{1.16}
\begin{longtable}{@{}>{\raggedright\arraybackslash}p{0.20\linewidth}>{\raggedright\arraybackslash}p{0.43\linewidth}>{\raggedright\arraybackslash}p{\dimexpr0.37\linewidth-16pt\relax}@{}}
\caption{Operational criteria for the 20 retained-behavior categories. Evidence must be present and enabled in the final submission. Examples identify observed mechanisms, not isolated causal effects.}\label{tab:behavior-criteria}\\
\toprule
Category & Evidence required for inclusion & Concrete retained example \\
\midrule
\endfirsthead
\multicolumn{3}{l}{\textit{Table~\thetable\ continued}}\\
\toprule
Category & Evidence required for inclusion & Concrete retained example \\
\midrule
\endhead
\midrule\multicolumn{3}{r}{\textit{Continued on next page}}\\
\endfoot
\bottomrule
\endlastfoot
Cache / memory capacity & A changed KV/cache allocation, reuse policy, cache representation, or memoized intermediate; an unchanged baseline cache is excluded. & Qwen on Inkling: larger GPU KV pool plus hierarchical host KV caching. \\
\addlinespace
Speculative decoding / MTP & Enabling or changing draft generation, verification depth, or native MTP/NGRAM settings. Baseline speculation alone is excluded. & DeepSeek on Qwen3.8: EAGLE steps/draft tokens change from 3/4 to 7/8. \\
\addlinespace
Graph / compilation & Newly enabled or retuned CUDA/NPU graph replay, capture sizes, or compiler execution paths. Profiling traces are not evidence of compilation. & Kimi on GLM-5.2: decode graph backend changes from disabled to full. \\
\addlinespace
Data movement / layout & A concrete reduction/reorganization of tensor copies, packing, transport, or layout on the active path; file packaging is excluded. & Astra on GLM-5.3 Flash: stride-aware paged-cache scatter kernels. \\
\addlinespace
Prefill / request scheduling & Changed chunk size, request admission, routing, batching of serving requests, or overlap scheduling; fixed evaluator concurrency is not changed. & GLM on VL GRPO: grouped five-sample rollouts and split-phase request processing. \\
\addlinespace
Attention backend & An explicit selection or dispatch change among existing attention implementations; an unchanged flag is excluded. & Kimi on Qwen3-VL: FA3 prefill with FlashInfer decode. \\
\addlinespace
Operator fusion & Combining previously separate operations/projections into fewer active kernels or joint computations; importing an already-used fused op is insufficient. & Claude on Qwen3-VL: fused norm, rotary, cache-write, and activation-quantization paths. \\
\addlinespace
Specialized GEMM / GEMV & A new/tuned matrix kernel, shape-specific dispatch, or matrix backend selected for an identified execution regime. & Astra on Qwen3-VL: small-batch FP8 GEMM; GLM on GLM-5.3: tuned MoE down-projection configs. \\
\addlinespace
Sparse / specialized attention & A changed attention computation or sparse-index path, beyond selecting an existing backend. & Astra on GLM-5.3: small-head BF16 sparse-MLA implementation; DeepSeek Flash: indexer top-k reuse. \\
\addlinespace
Host runtime overhead & Reduced Python dispatch, host synchronization, polling, garbage collection, or thread/socket overhead. Startup-only repairs are excluded. & DeepSeek on MiMo SFT: shorter API/engine future-poll intervals. \\
\addlinespace
Parallel topology / layout & Changed TP/PP/EP, replica placement, sequence/context parallelism, or distribution of model work across the allocated devices. & Astra and Seed on Kimi: TP8/PP4 replaces TP32 across four nodes. \\
\addlinespace
Input preprocessing & Changed image/audio/datum decoding, hashing, token/tensor construction, or preprocessing parallelism. Workload/data substitutions are excluded. & Kimi on VL GRPO: parallel image/datum preprocessing; Claude on action inference: fast image-array JSON decoding. \\
\addlinespace
Communication / overlap & Changed collective execution/protocol, weight-transfer scheduling, or explicit stream overlap. A network-connectivity repair alone is excluded. & Astra on DeepSeek inference: FlashComm; GLM on LingBot: context-update/VAE overlap. \\
\addlinespace
Model-specific pipeline & Reorganized VAE, codec, denoising, or media stages while retaining an identifiable model-specific execution change. & Kimi on Wan: temporal VAE batching with unchanged spatial tiling. \\
\addlinespace
Quantization / weight layout & Changed weight/KV representation, dequantization, packing, or weight-layout dispatch; baseline checkpoint precision alone is excluded. & Astra on Nemotron: NVFP4 dequantization; Qwen on Muse: NN weight layout for decode. \\
\addlinespace
Training memory / state & Changed recomputation, parameter/optimizer residency, allocator lifecycle, or training-state transfer. Checkpoint-path repairs are excluded. & Claude on MiMo SFT: selective recomputation and allocator reuse; Kimi VL GRPO: resident reference parameters. \\
\addlinespace
Sequence-state path & Changed recurrent-state/causal-convolution update implementation, beyond generic cache capacity. & GLM on GLM-5.3 Flash: fused KDA causal-convolution path (formal zero). \\
\addlinespace
Batch / microbatch tuning & Changed training/forward microbatch token budgets or packing; the prescribed global batch and update counts remain separately checked. & Kimi on MiMo SFT: microbatch token cap changes from 2048 to 768. \\
\addlinespace
Precision / accumulation & Changed active arithmetic, rounding, or deterministic-kernel policy; claims of numerical equivalence do not replace the quality gate. & Kimi on Nemotron: deterministic FP16 state rounding; Qwen on MiniMax: FP8 attention. \\
\addlinespace
Work reduction / reuse & Explicitly omitted computation or reused an equivalent result. Record the behavior even when the evaluator rejects its semantics. & Claude on Wan: reuse identical CFG attention; Seed TTS: remove repetition-penalty processing (compliance rejection). \\
\end{longtable}
\endgroup

\end{document}